\documentclass[runningheads]{llncs}

\usepackage{graphicx}
\usepackage{comment}
\usepackage{amsmath,amssymb}
\usepackage{color}
\usepackage{url}
\usepackage{hyperref}
\usepackage{algorithm,algpseudocode}
\usepackage{subcaption}

\usepackage{stmaryrd}
\usepackage{stackrel}
\usepackage[english]{babel}
\graphicspath{{Figures/}}

\newif\ifreview
\reviewfalse

\ifreview
	\usepackage{lineno}

	\linenumbers
\fi

\begin{document}

\def\SubNumber{100}

\def\GCPRTrack{Regular Track}

\title{Multidirectional Conjugate Gradients for Scalable Bundle Adjustment}

\ifreview
\else

	\author{Simon Weber \and
	Nikolaus Demmel \and
	Daniel Cremers}
	
	\authorrunning{S. Weber et al.}
	
	\institute{Technical University of Munich\\ 
	\email{\{sim.weber,nikolaus.demmel,cremers\}@tum.de}}
\fi

\maketitle              

\begin{abstract}
We revisit the problem of large-scale bundle adjustment and propose a technique called Multidirectional Conjugate Gradients that accelerates the solution of the normal equation by up to 61\%. The key idea is that we enlarge the search space of classical preconditioned conjugate gradients to include multiple search directions. As a consequence, the resulting algorithm requires fewer iterations, leading to a significant speedup of large-scale reconstruction, in particular for denser problems where traditional approaches notoriously struggle. We provide a number of experimental ablation studies revealing the robustness to variations in the hyper-parameters and the speedup as a function of problem density.
\end{abstract}

\section{Introduction}

The classical challenge of image-based large scale reconstruction is witnessing renewed interest with the emergence of large-scale internet photo collections \cite{key-15}.  The computational bottleneck of 3D reconstruction and structure from motion methods is the problem of large-scale bundle adjustment (BA): Given a set of measured image feature locations and correspondences, BA aims to jointly estimate the 3D landmark positions and camera parameters by minimizing a non-linear least squares reprojection error. More specifically, the most time-consuming step is the solution of the normal equation in the popular Levenberg-Marquardt (LM) algorithm that is typically solved by Preconditioned Conjugate Gradients (PCG).

In this paper, we propose a new iterative solver for the normal equation that relies on the decomposable structure of the competitive block Jacobi preconditioner. Inspired by respective approaches in the domain-decomposition literature, we exploit the specificities of the Schur complement matrix to enlarge the search-space of the traditional PCG approach leading to what we call {\em Multidirectional Conjugate Gradients} (MCG). In particular our contributions are as follows:
\begin{figure}[ht!]
\begin{subfigure}{0.5\textwidth}
\includegraphics[width=0.90\linewidth, height=5cm]{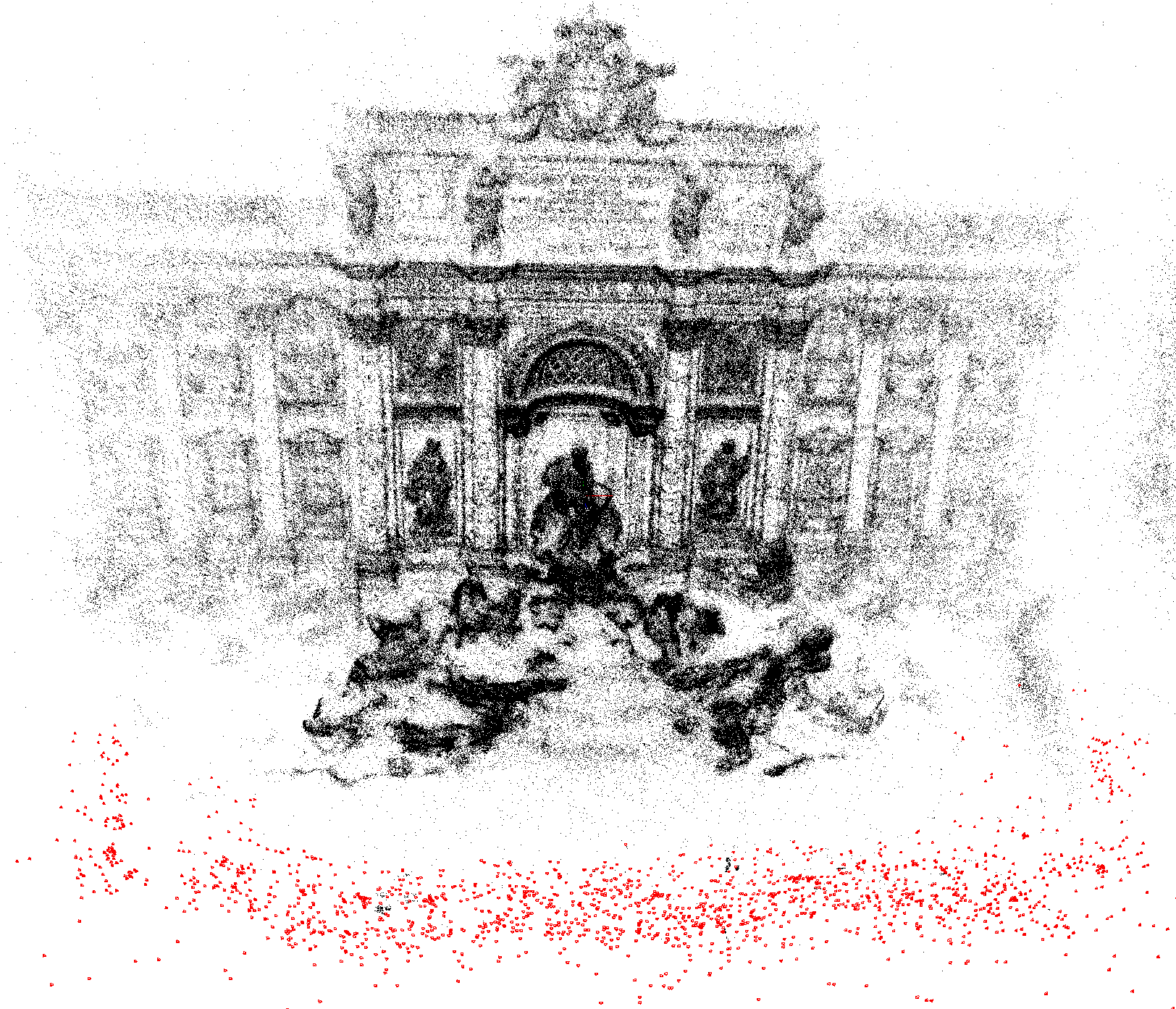} 
\caption{\textit{Final-1936} from BAL dataset}
\label{fig:subim1}
\end{subfigure}
\begin{subfigure}{0.5\textwidth}
\includegraphics[width=0.90\linewidth, height=5cm]{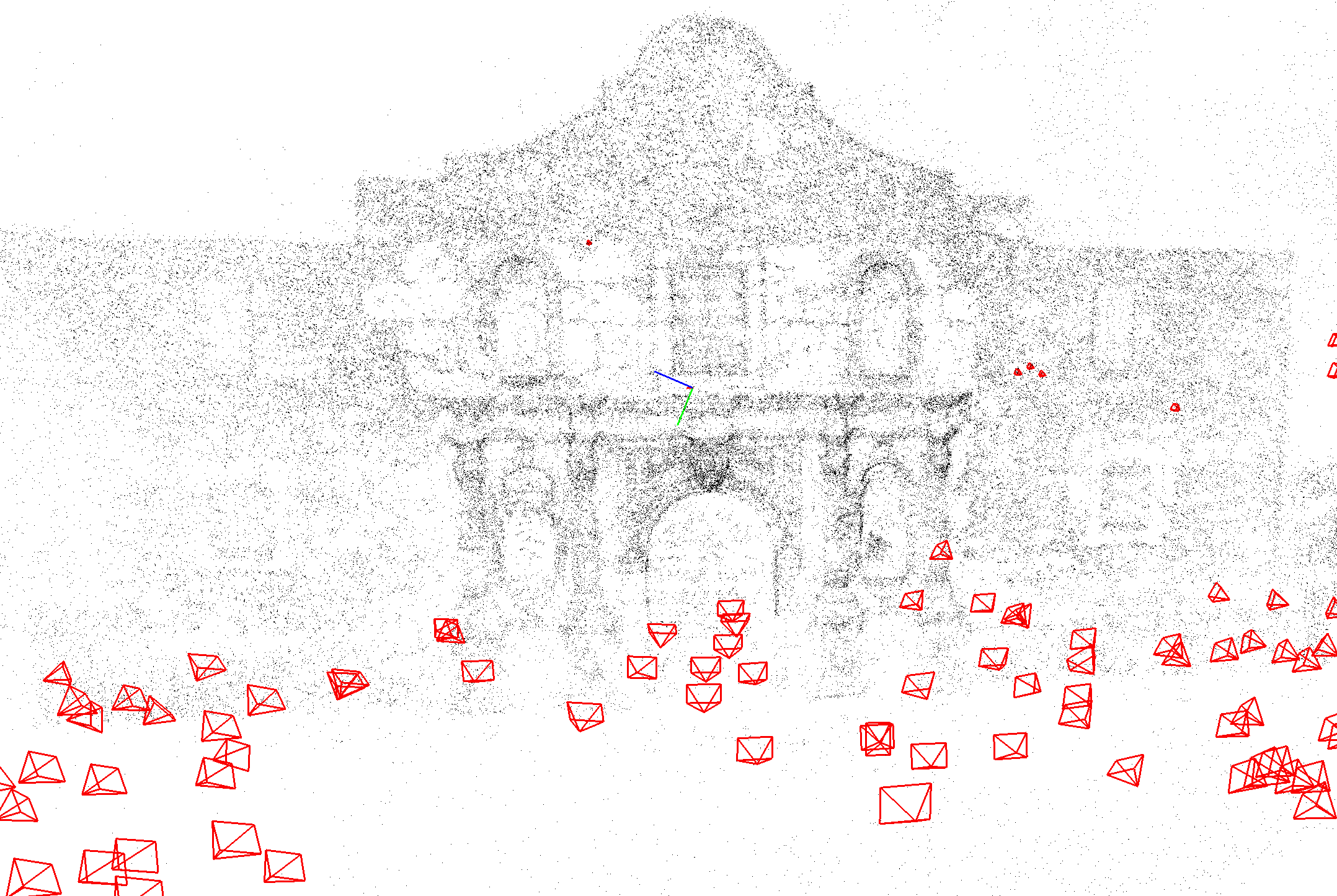}
\caption{\textit{Alamo} from 1dSfM dataset}
\label{fig:subim2}
\end{subfigure}
\caption{(a) Optimized 3D reconstruction of a \textit{final} BAL dataset with $1936$ poses and more than five million observations. For this problem MCG is \text{39\%} faster than PCG and the overall BA resolution is \text{16\%} faster. (b) Optimized 3D reconstruction of Alamo dataset from 1dSfM with $571$ poses and $900000$ observations. For this problem MCG is \text{56\%} faster than PCG and the overall BA resolution is \text{22\%} faster.}
\label{fig:image2}
\end{figure}
\begin{itemize}
    \item[$\bullet$] We design an extension of the popular PCG by using local contributions of the poses to augment the space in which a solution is sought for.
    \item[$\bullet$] We experimentally demonstrate the robustness of MCG with respect to the relevant hyper-parameters.
    \item[$\bullet$] We evaluate MCG on a multitude of BA problems from BAL \cite{key-1} and 1dSfM \cite{wilson2014eccv_1dsfm} datasets with different sizes and show that it is a promising alternative to PCG.
    \item[$\bullet$] We experimentally confirm that the performance gain of our method increases with the density of the Schur complement matrix leading to a speedup for solving the normal equation of up to 61\%.
\end{itemize}

\section{Related Work}

Since we propose a way to solve medium to large-scale BA using a new iterative
solver that enlarges the search-space of the traditional
PCG, in the following we will review both scalable BA and recent CG literature.

\subsection*{Scalable bundle adjustment}

A detailed survey of the theory and methods in BA literature can be
found in \cite{key-2}. Sparsity of the BA problem is commonly exploited
with the Schur complement matrix \cite{key-14}. As the performance
of BA methods is closely linked to the resolution of the normal equations,
speed up the solve step is a challenging task. Traditional direct
solvers such as sparse or dense Cholesky factorization \cite{key-11}
have been outperformed by inexact solvers as the problem size increases
and are therefore frequently replaced by Conjugate Gradients (CG) based methods 
\cite{key-1,key-12,key-13}. As its convergence
rate depends on the condition number of the linear system a preconditioner
is used to correct ill-conditioned BA problems \cite{key-10}. Several works
tackle the design of performant preconditioners for BA:
\cite{key-3} proposed the band block diagonals of the Schur complement
matrix, \cite{key-4} exploited the strength of the coupling between
two poses to construct cluster-Jacobi and block-tridiagonal preconditioners,
\cite{key-5} built on the combinatorial structure of BA. However, despite these advances in the design of preconditioners, the iterative solver itself has rarely been challenged.

\subsection*{(Multi-preconditioned) conjugate gradients}

Although CG has been a popular iterative solver for decades \cite{key-6} there exist some interesting recent innovations, 
e.g. flexible methods with a preconditioner that changes throughout the iteration \cite{key-17}.
The case of a preconditioner that can be decomposed into a sum of preconditioners has been exploited by using Multi-Preconditioned Conjugate Gradients
(MPCG) \cite{key-7}. Unfortunately, with increasing system size MPCG rapidly becomes inefficient. As a remedy, Adaptive Multi-Preconditioned Conjugate Gradients have recently been proposed \cite{key-8,key-20}. This approach is particularly well adapted for domain-decomposable problems \cite{key-9}. While decomposition of the reduced camera system in BA has already been tackled e.g. with stochastic clusters in \cite{key-18}, to our knowledge the decomposition {\em inside} the iterative solver has never been explored. As we will show in the following, this modification gives rise to a significant boost in performance.

\section{Bundle Adjustment and Multidirectional Conjugate Gradients}

We consider the general form of bundle adjustment with $n_{p}$ poses and $n_{l}$ landmarks. Let $x$ be the state
vector containing all the optimization variables. It is divided into
a pose part $x_{p}$ of length $d_{p}n_{p}$ containing extrinsic
and eventually intrinsic camera parameters for all poses (generally
$d_{p}=6$ if only extrinsic parameters are unknown and $d_{p}=9$
if intrinsic parameters also need to be estimated) and a landmark
part $x_{l}$ of length $3n_{l}$ containing the 3D coordinates of
all landmarks. Let $r\left(x\right)=[r_{1}\left(x\right),...,r_{k}\left(x\right)]$ 
be the vector of residuals for a 3D reconstruction. The objective
is to minimize the sum of squared residuals 
\begin{equation}
F\left(x\right)=\lVert r\left(x\right)\rVert^{2} = \sum_i \lVert r_i(x) \rVert^{2}
\end{equation}

\subsection{Least Squares Problem and Schur Complement}

This minimization problem is usually solved with the Levenberg 
Marquardt algorithm, which is based on the first-order Taylor approximation
of $r\left(x\right)$ around the current state estimate $x^{0}=\left(x_{p}^{0},x_{l}^{0}\right)$:

\begin{align}
r\left(x\right) & \approx r^{0}+J\Delta x
\end{align}
 where 
 
\begin{align}
r^{0} & =r\left(x^{0}\right),\\
\Delta x & = x-x^{0},\\
J & =\frac{\partial r}{\partial x}\mid_{x=x^{0}}
\end{align}
and $J$ is the Jacobian of $r$ that is decomposed into a pose part
$J_{p}$ and a landmark part $J_{l}$.
An added regularization term that improves convergence gives the damped
linear least squares problem 

\begin{align}
\underset{\Delta x_{p},\Delta x_{l}}{\text{min}} & \left(\lVert r^{0}+\left(\begin{array}{cc}
J_{p} & J_{l}\end{array}\right)\left(\begin{array}{c}
\Delta x_{p}\\
\Delta x_{l}
\end{array}\right)\rVert^{2}+\lambda\lVert\left(\begin{array}{cc}
D_{p} & D_{l}\end{array}\right)\left(\begin{array}{c}
\Delta x_{p}\\
\Delta x_{l}
\end{array}\right)\rVert^{2}\right)
\end{align}
 with $\lambda$ a damping coefficient and $D_{p}$ and $D_{c}$ diagonal
damping matrices for pose and landmark variables. 
This damped problem leads to the corresponding normal equation 

\begin{align}
H\left(\begin{array}{c}
\Delta x_{p}\\
\Delta x_{l}
\end{array}\right) & =-\left(\begin{array}{c}
b_{p}\\
b_{l}
\end{array}\right)
\end{align}
 where 
 
\begin{align}
H & = \left(\begin{array}{cc}
U_{\lambda} & W\\
W^{\top} & V_{\lambda}
\end{array}\right),\\
U_{\lambda} & =J_{p}^{\top}J_{p}+\lambda D_{p}^{\top}D_{p},\\
V_{\lambda} & =J_{l}^{\top}J_{l}+\lambda D_{l}^{\top}D_{l},\\
W & =J_{p}^{\top}J_{l},\text{ }b_{p}=J_{p}^{\top}r^{0},\\
b_{l} & =J_{l}^{\top}r^{0}
\end{align}
As the system matrix $H$ is of size $\left(d_{p}n_{p}+3n_{l}\right)^{2}$
and tends to be excessively costly for large-scale problems \cite{key-1},
it is common to reduce it by using the Schur complement trick and
forming the reduced camera system 

\begin{align}
S\Delta x_{p} & =-\widetilde{b}
\end{align}
 with 
 
\begin{align}
S & =U_{\lambda}-WV_{\lambda}^{-1}W^{\top},\\
\widetilde{b} & =b_{p}-WV_{\lambda}^{-1}b_{l}
\end{align}
and then solving (13) for $\Delta x_{p}$ and backsubstituting
$\Delta x_{p}$ in 

\begin{align}
\Delta x_{l} & =-V_{\lambda}^{-1}\left(-b_{l}+W^{\top}\Delta x_{p}\right)
\end{align}

\subsection{Multidirectional Conjugate Gradients}

Direct methods such as Cholesky decomposition \cite{key-2} have been
studied for solving (13) for small-size problems, but this approach
implies a high computational cost whenever problems become too large.

A very popular iterative solver for large symmetric positive-definite
system is the CG algorithm \cite{key-16}. Since its
convergence rate depends on the distribution of eigenvalues of $S$
it is common to replace (13) by a preconditioned system. Given a preconditioner
$M$ the preconditioned linear system associated to 

\begin{align}
S\Delta x_{p}=-\widetilde{b}
\end{align}
is 

\begin{align}
M^{-1}S\Delta x_{p}=-M^{-1}\widetilde{b}
\end{align}
and the resulting algorithm
is called Preconditioned Conjugate Gradients (PCG) (see Algorithm 1).
For block structured matrices as $S$ a competitive preconditioner
is the block diagonal matrix $D\left(S\right)$, also called block Jacobi
preconditioner \cite{key-1}. It is composed of the block diagonal elements of $S$. 
Since the block $S_{mj}$ of $S$ is nonzero
if and only if cameras $m$ and $j$ share at least one common point,
each diagonal block depends on a unique pose and is applied to the
part of conjugate gradients residual $r_{i}^{j}$ that is associated
to this pose. The motivation of this section is to enlarge the conjugate
gradients search space by using several local contributions instead
of a unique global contribution.

\begin{algorithm}
\caption{Preconditioned Conjugate Gradients}\label{algo}
\begin{algorithmic}[1]
\State $x_{0}$,~$r_{0}=-\widetilde{b}-Sx_{0}$,~$Z_{0}=D(S)^{-1}r_{0}$,~$P_{0}=Z_{0}$,~$\epsilon$;
\While{$i<\text{imax}$}
    \State $Q_{i}=SP_{i}$;
    \State $\Delta_{i}=Q_{i}^{\top}P_{i}$;~$\gamma_{i}=P_{i}^{\top}r_{i}$;~$\alpha_{i}=\frac{\gamma_{i}}{\Delta_{i}}$;
    \State $x_{i+1}=x_{i}+\alpha_{i}P_{i}$;
    \State $r_{i+1}=r_{i}-\alpha_{i}Q_{i}$;
    \If{$r_{i+1}<\epsilon * r_{0}$}
        \State break
    \EndIf
    \State $Z_{i+1}=D(S)^{-1}r_{i+1}$;
    \State $\Phi_{i}=Q_{i}^{\top}Z_{i+1}$;~$\beta_{i}=\frac{\Phi_{i}}{\Delta_{i}}$;
    \State $P_{i+1}=Z_{i+1}-\beta_{i}P_{i}$;
\EndWhile
\State return~$x_{i+1}$;
\end{algorithmic}
\end{algorithm}

\bigskip
\subsubsection{Adaptive multidirections}

\begin{figure}[ht!]
\begin{subfigure}{0.6\textwidth}
\begin{center}
\includegraphics[width=0.8\linewidth, height=2.4cm]{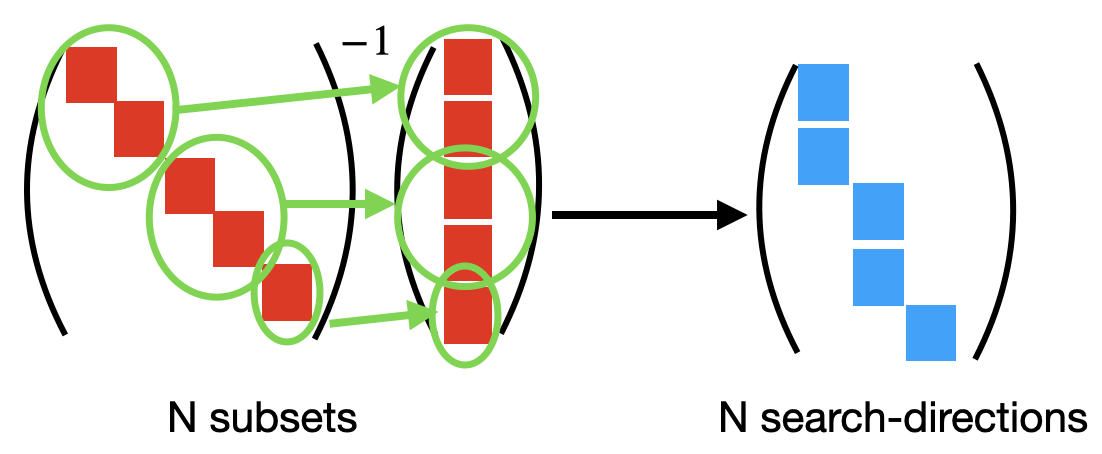} 
\end{center}
\caption{Decomposed preconditioned CG residuals}
\label{fig:subim1}
\end{subfigure}
\begin{subfigure}{0.4\textwidth}
\begin{center}
\includegraphics[width=0.8\linewidth, height=2.5cm]{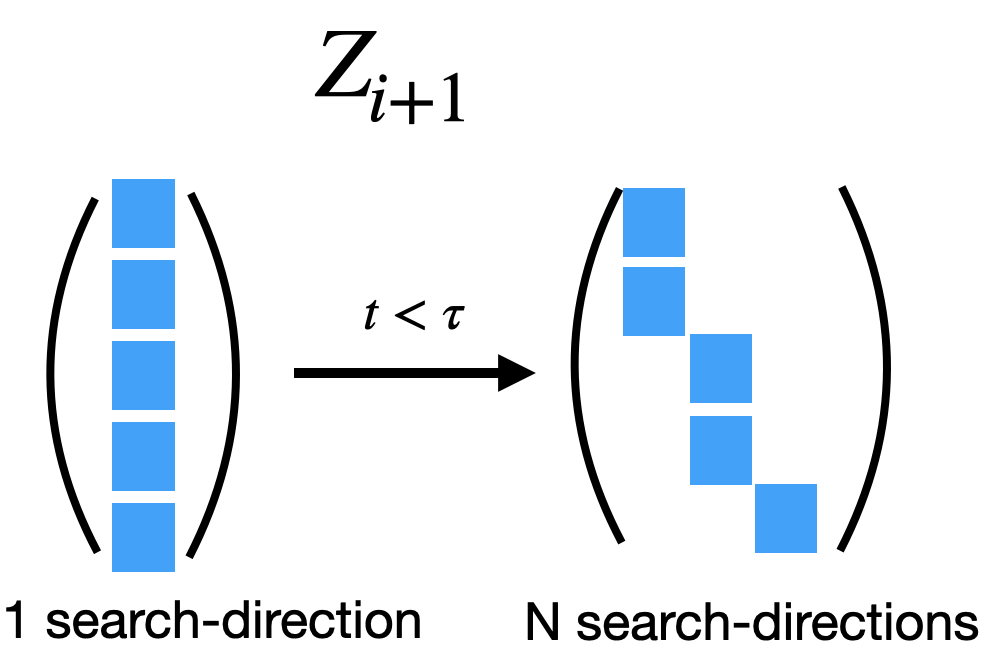}
\end{center}
\caption{Enlarged search-space}
\label{fig:subim2}
\end{subfigure}
\caption{(a) Block-Jacobi preconditioner $D\left(S\right)$ is divided into $N$ submatrices $D_{p}\left(S\right)$ and each of them is directly applied to the associated block-row $r^{p}$ in the CG residual. (b) Up to a $\tau$-test the search-space is enlarged. Each iteration provides $N$ times more search-directions than PCG.}
\label{fig:image2}
\end{figure}

\paragraph{Local preconditioners.}

We propose to decompose the set of poses into $N$ subsets of sizes
$l_{1}$,...,$l_{N}$ and to take into consideration the block-diagonal
matrix $D_{p}\left(S\right)$ of the block-jacobi preconditioner and
the associated residual $r^{p}$ that correspond to the $l_{p}$ poses
of subset $p$ (see Figure 2(a)). All direct solves are performed inside these subsets and not in the global set.
Each local solve is treated as a separate preconditioned equation
and provides a unique search-direction. Consequently the conjugate
vectors $Z_{i+1}\in\mathbb{R}^{d_{p}n_{p}}$ in the preconditioned conjugate
gradients (line 10 in Algorithm 1) are now replaced by conjugate matrices
$Z_{i+1}\in\mathbb{R}^{d_{p}n_{p}\times N}$ whose each column
corresponds to a local preconditioned solve. The search-space is then
significantly enlarged: $N$ search directions are generated at each
inner iteration instead of only one. An important drawback is that
matrix-vector products are replaced by matrix-matrix products which
can lead to a significant additional cost. A trade-off between convergence
improvement and computational cost needs to be designed.

\begin{algorithm}
\caption{Multidirectional Conjugate Gradients}\label{algo}
\begin{algorithmic}[1]
\State $x_{0}$,~$r_{0}=-\widetilde{b}-Sx_{0}$,~$Z_{0}=D(S)^{-1}r_{0}$,~$P_{0}=Z_{0}$,~$\epsilon$;
\While{$i<\text{imax}$}
    \State $Q_{i}=SP_{i}$;
    \State $\Delta_{i}=Q_{i}^{\top}P_{i}$;~$\gamma_{i}=P_{i}^{\top}r_{i}$;~$\alpha_{i}=\Delta_{i}^{\dagger}\gamma_{i}$;
    \State $x_{i+1}=x_{i}+P_{i}\alpha_{i}$;
    \State $r_{i+1}=r_{i}-Q_{i}\alpha_{i}$;
    \If{$r_{i+1}<\epsilon * r_{0}$}
        \State break
    \EndIf
    \State $t_{i}=\frac{\gamma_{i}^{\top}\alpha_{i}}{r_{i+1}^{\top}D(S)^{-1}r_{i+1}}$;
    \If{$t_{i}<\tau$}
        \State $Z_{i+1}=\left(\begin{array}{c}
                D_{1}(S)^{-1}r_{i+1}^{1}\\
                0\\
                ...\\
                0
                \end{array}\begin{array}{c}
                \\
                ...\\
                ...\\
                \\
                \end{array}\begin{array}{c}
                0\\
                ...\\
                0\\
                D_{N}(S)^{-1}r_{i+1}^{N}
                \end{array}\right)$;
    \Else
        \State $Z_{i+1}=D(S)^{-1}r_{i+1}$;
    \EndIf
    \State $\Phi_{i,j}=Q_{j}^{\top}Z_{i+1}$;~$\beta_{i,j}=\Delta_{j}^{\dagger}\Phi_{i,j}$~for~$j=0,...,i$;
    \State $P_{i+1}=Z_{i+1}-\sum_{j=0}^{i}P_{j}\beta_{i,j}$;
\EndWhile
\State return~$x_{i+1}$;
\end{algorithmic}
\end{algorithm}

\paragraph{Adaptive $\tau$-test.}

Following a similar approach as in \cite{key-8} we propose to use
an adaptive multidirectional conjugate gradients algorithm (MCG, see
Algorithm 2) that adapts automatically if the convergence is too slow.
Given a threshold $\tau\in\mathbb{R}^{+}$ chosen by the user, a $\tau$-test
determines whether the algorithm sufficiently reduces the error (case
$t_{i}>\tau$) or not (case $t_{i}<\tau$). In the first case a global
block Jacobi preconditioner is used and the algorithm performs a step
of PCG; in the second case local block Jacobi preconditioners are
used and the search-space is enlarged (see Figure 2(b)).

\subsubsection{Optimized implementation}

Besides matrix-matrix products two other changes appear. Firstly an $N\times N$ matrix $\Delta_{i}$
must be inverted (or pseudo-inverted if $\Delta_{i}$ is not full-rank)
each time $t_{i}<\tau$ (line 4 in Algorithm 2). Secondly a full reorthogonalization
is now necessary (line 16 in Algorithm 2) because of numerical errors
while $\beta_{i,j}=0$ as soon as $i\neq j$ in PCG. 

To improve the efficiency of MCG we do not directly
apply $S$ to $P_{i}$ (line 3 in Algorithm 2) when the search-space
is enlarged. By construction the block $S_{kj}$ is nonzero if and
only if cameras $k$ and $j$ observe at least one common point. The
trick is to use the construction of $Z_{i}$ and to directly apply
the non-zero blocks $S_{jk}$, i.e. consider only poses $j$ observing
a common point with $k$, to the column in $Z_{i}$ associated to
the subset containing pose $k$ and then to compute 

\begin{align}
Q_{i} & =SZ_{i}-\stackrel[j=0]{i-1}{\sum}Q_{j}\beta_{i,j}
\end{align}

To get $t_{i}$ we need to use a global solve (line 10 in Algorithm
2). As the local block Jacobi preconditioners $\{D_{p}(S)\}_{p=1,...,N}$
and the global block Jacobi preconditioner $D\left(S\right)$ share
the same blocks it is not necessary to derive all local solves to
construct the conjugates matrix (line 12 in Algorithm 2); instead it
is more efficient to fill this matrix with block-row elements of the
preconditioned residual $D\left(S\right)^{-1}r_{i+1}$.

As the behaviour of CG residuals is \textit{a priori} unknown the best decomposition is not obvious. We decompose the set of poses into $N-1$ subsets of same size and the last subset is filled by the few remaining poses. This structure presents the practical advantage to be very easily fashionable and the parallelizable block operations are balanced.

\section{Experimental Evaluations}

\subsection{Algorithm and Datasets}

\paragraph*{Levenberg-Marquardt (LM) loop.}

Starting with damped parameter $10^{-4}$ we update $\lambda$ according
to the success or failure of the LM loop. Our implementation runs
for at most $25$ iterations, terminating early if a relative function
tolerance of $10^{-6}$ is reached. Our evaluation is built on the LM loop implemented in \cite{key-18} and we also estimate intrinsics parameters for each pose. 

\paragraph*{Iterative solver step.}

For a direct performance comparison we implement our own MCG and PCG solvers in C++ by using Eigen 3.3 library. All row-major-sparse matrix-vector and matrix-matrix products are multi-threaded by using $4$ cores. The tolerance $\epsilon$ and the maximum number of iterations are set to $10^{-6}$ and $1000$ respectively. Pseudo-inversion is derived with the pseudo-inverse function from Eigen.

\paragraph*{Datasets.}

For our evaluation we use $9$ datasets with different sizes and heterogeneous Schur complement
matrix densities $d$ from BAL \cite{key-1} and 1dSfM \cite{wilson2014eccv_1dsfm} datasets (see Table 1). The values of $N$ and $\tau$ are arbitrarily chosen and the robustness of our algorithm to these parameters is discussed in the next subsection.  

\begin{table}
\caption{Details of the problems from BAL (prefixed as: F for \textit{final}, L for \textit{Ladybug}) and 1dSfM used in our experiments. $d$ is the density of the associated Schur complement matrix, $N$ is the number of subsets, $\tau$ is the adaptive threshold that enlarges the search-direction space.} \label{tab1} \centering
\begin{tabular}{|c|c|c|c|c|c|c|}
\hline
Names &  poses & points & projections & $d$ & $N$ & $\tau$\\
\hline
Piazza del Popolo & 335 & 37,609 & 195,016 & 0.57 & 33 & 10\\
Metropolis & 346 & 55,679 & 255,987 & 0.50 & 34 & 6\\
F-394 & 394 & 100,368 & 534,408 & 0.94 & 131 & 6\\
Montreal & 459 & 158,005 & 860,116 & 0.60 & 22 & 3\\
Notre-Dame & 547 & 273,590 & 1,534,747 & 0.77 & 45 & 10\\
Alamo & 571 & 151,085 & 891,301 & 0.77 & 47 & 10\\
L-646 & 646 & 73,484 & 327,297 & 0.25 & 64 & 2\\
F-871 & 871 & 527,480 & 2,785,977 & 0.40 & 43 & 2.5\\
F-1936 & 1,936 & 649,673 & 5,213,733 & 0.91 & 121 & 3.3 \\
\hline
\end{tabular}
\end{table}

We run experiments on MacOS 11.2 with Intel Core i5 and $4$ cores at 2GHz. 

\subsection{Sensitivity with $\tau$ and $N$}

In this subsection we are interested in the solver runtime ratio that is defined as $\frac{t_{MCG}}{t_{PCG}}$ where $t_{MCG}$ (resp. $t_{PCG}$) is the total runtime to solve all the linear systems (12) with MCG (resp. PCG) until a given BA problem converges. We investigate the influence of $\tau$ and $N$ on this ratio.

\paragraph*{Sensitivity with $\tau$.}

We solve BA problem for different values of $\tau$ and for a fixed number of subsets $N$ given in Table 1.
For each problem a wide range of values supplies a good trade-off between
the augmented search-space and the additional computational cost (see Figure 3).
Although the choice of $\tau$ is crucial it does
not require a high accuracy. That confirms the tractability of our
solver with $\tau$.

\begin{figure}[ht!]
\includegraphics[width=\textwidth]{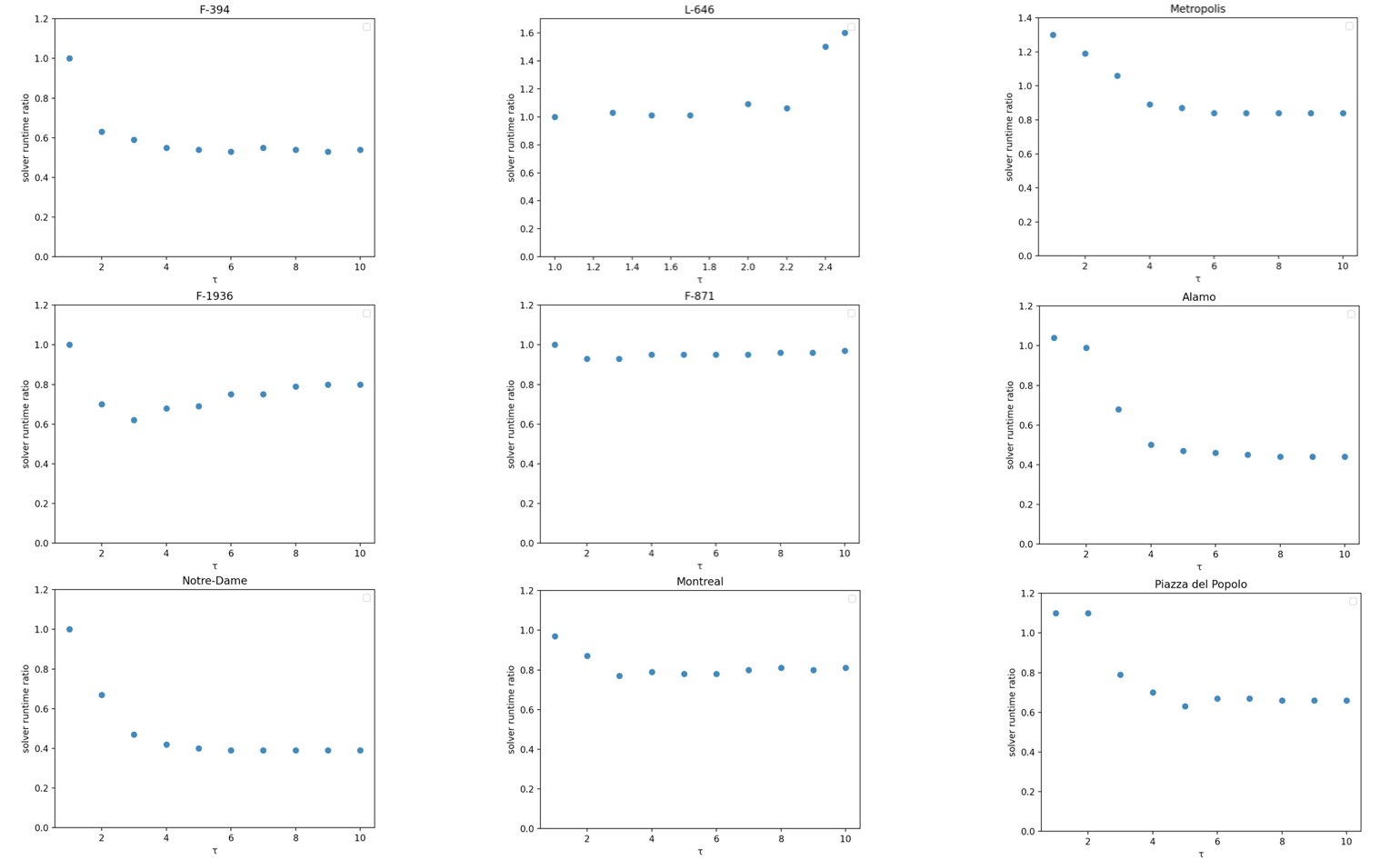}
\caption{Robustness to $\tau$. The plots show the performance ratio as a function of $\tau$ for a number of subsets given in Table 1. The wide range of values that give similar performance confirms the tractability of MCG with $\tau$.} \label{fig1}
\end{figure}

\paragraph*{Sensitivity with $N$.}

Similarly we solve BA problem for different values of $N$ and for a fixed $\tau$ given in Table 1.
For each problem a wide range of values supplies a good trade-off between
the augmented search-space and the additional computational cost (see Figure 4). That confirms the tractability
of our solver with $N$. 

\begin{figure}[ht!]
\includegraphics[width=\textwidth]{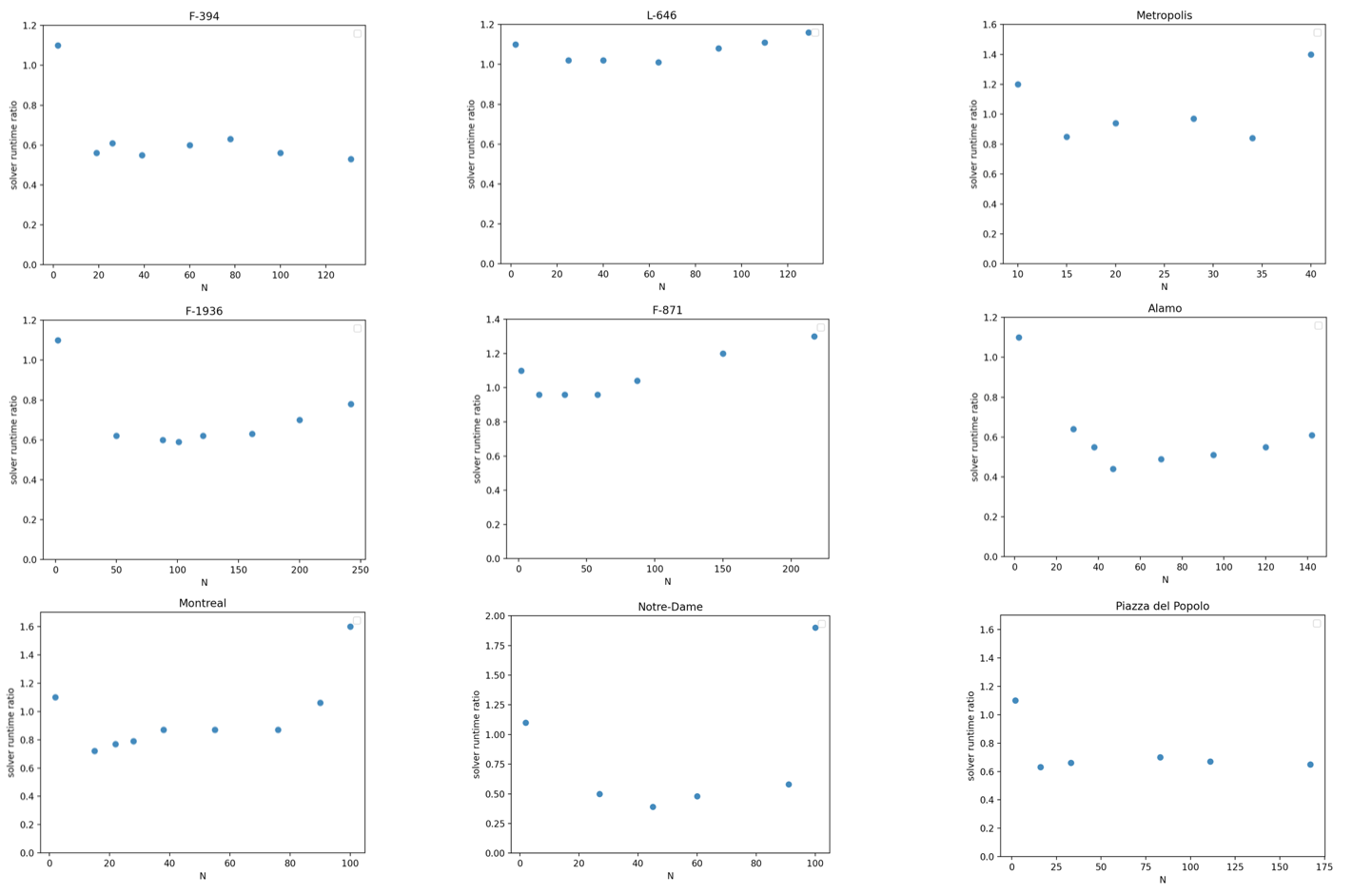}
\caption{Robustness to the number of subsets $N$. The plots show the performance ratio as a function of $N$ for $\tau$ given in Table 1. The wide range of values that give similar performance confirms the tractability of MCG with $N$.} \label{fig1}
\end{figure}

\begin{figure}[ht!]
\begin{center}
\includegraphics[width=0.6\textwidth]{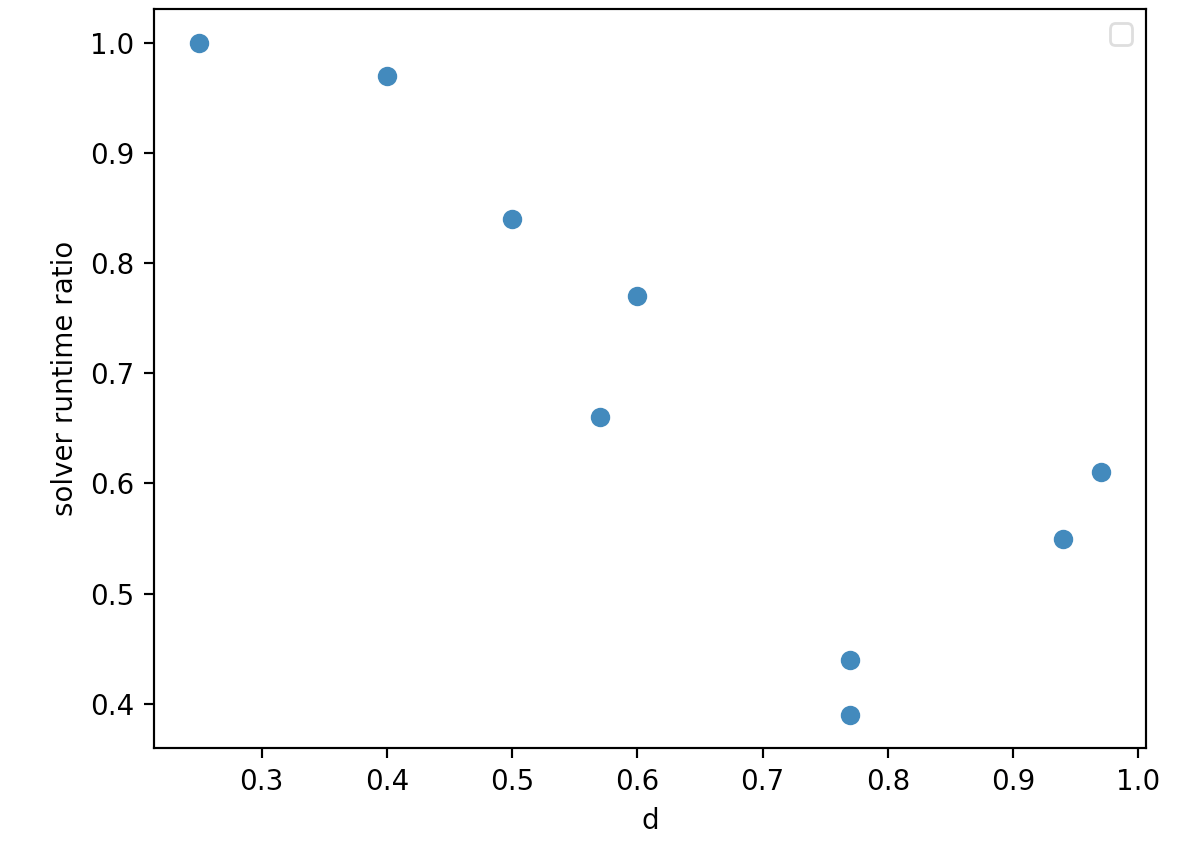}
\caption{Density effect on the relative performance. Each point represents a BA problem from Table 1 and $d$ is the density of the Schur complement matrix. Our solver competes PCG for sparse Schur matrix and leads to a significant speed-up for dense Schur matrix.} \label{fig1}
\end{center}
\end{figure}

\subsection{Density effect}

As the performance of PCG and MCG depends on matrix-vector product
and matrix-matrix product respectively we expect a correlation with
the density of the Schur matrix. Figure 5 investigates this intuition: MCG greatly outperfoms PCG for dense Schur matrix and
is competitive for sparse Schur matrix.

\subsection{Global performance}

Figures 6 and 7 present the total runtime with respect to the number of BA iterations for each problem and the convergence plots of total BA cost for \textit{F-1936} and \textit{Alamo} datasets, respectively. MCG and PCG give the same error at each BA iteration but the first one is more efficient in terms of runtime. Table 2 summarizes our results and highlights the great performance of MCG for dense Schur matrices. In the best case BA resolution is more than $20\%$ faster than using PCG. Even for sparser matrices MCG competes PCG: in the worst case MCG presents similar results as PCG. If we restrict our comparison to the linear system solve steps our relative results are even better: MCG is up to $60\%$ faster than PCG and presents similar results as PCG in the worst case.

\begin{figure}[ht!]
\includegraphics[width=\textwidth]{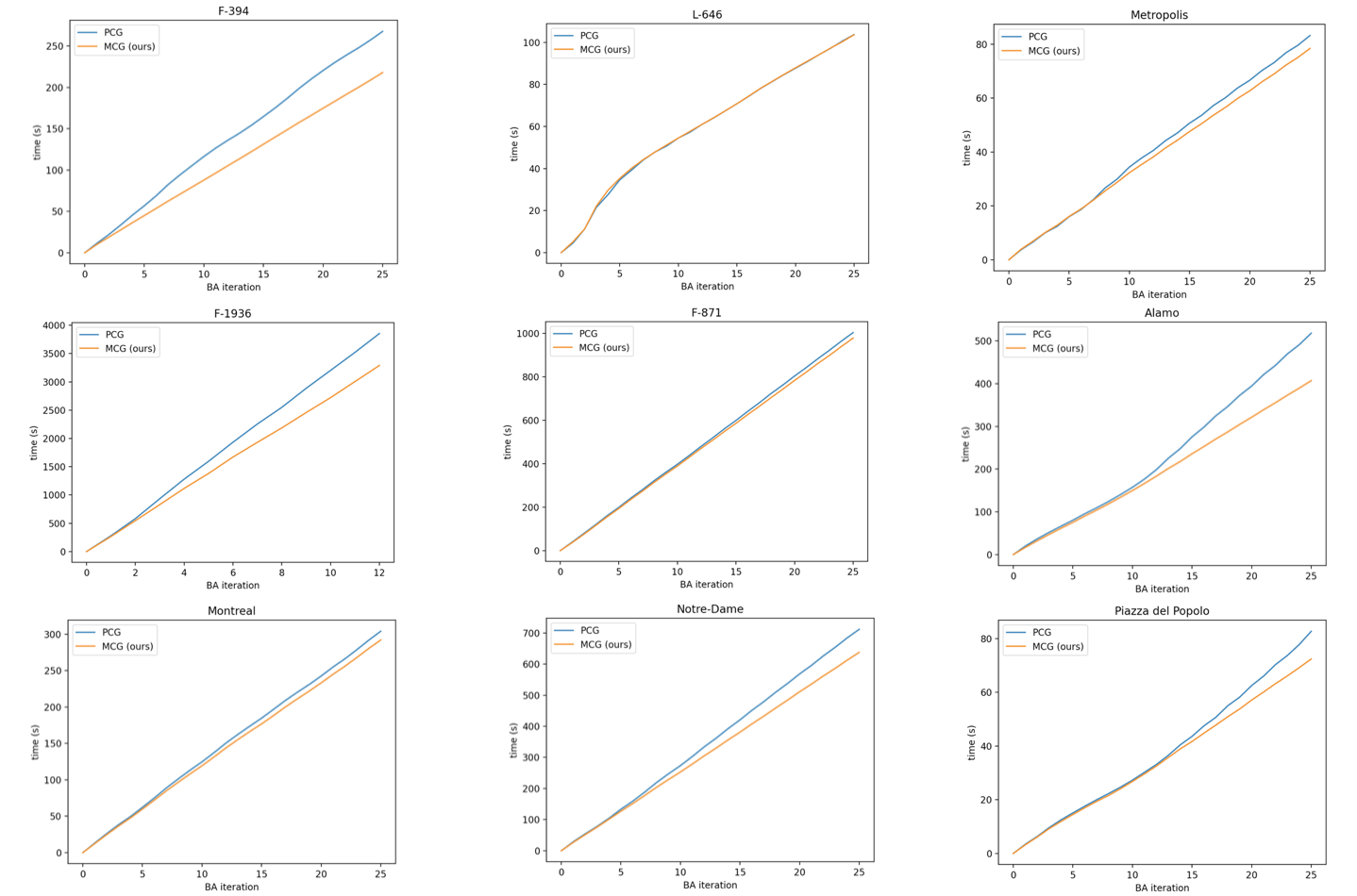}
\caption{Global runtime to solve BA problem. The plots represent the total time with respect to the number of BA iterations. For almost all problems the BA resolution with MCG (orange) is significantly faster than PCG (blue).} \label{fig1}
\end{figure}

\begin{figure}[ht!]
\begin{subfigure}{0.5\textwidth}
\includegraphics[width=0.90\linewidth, height=5cm]{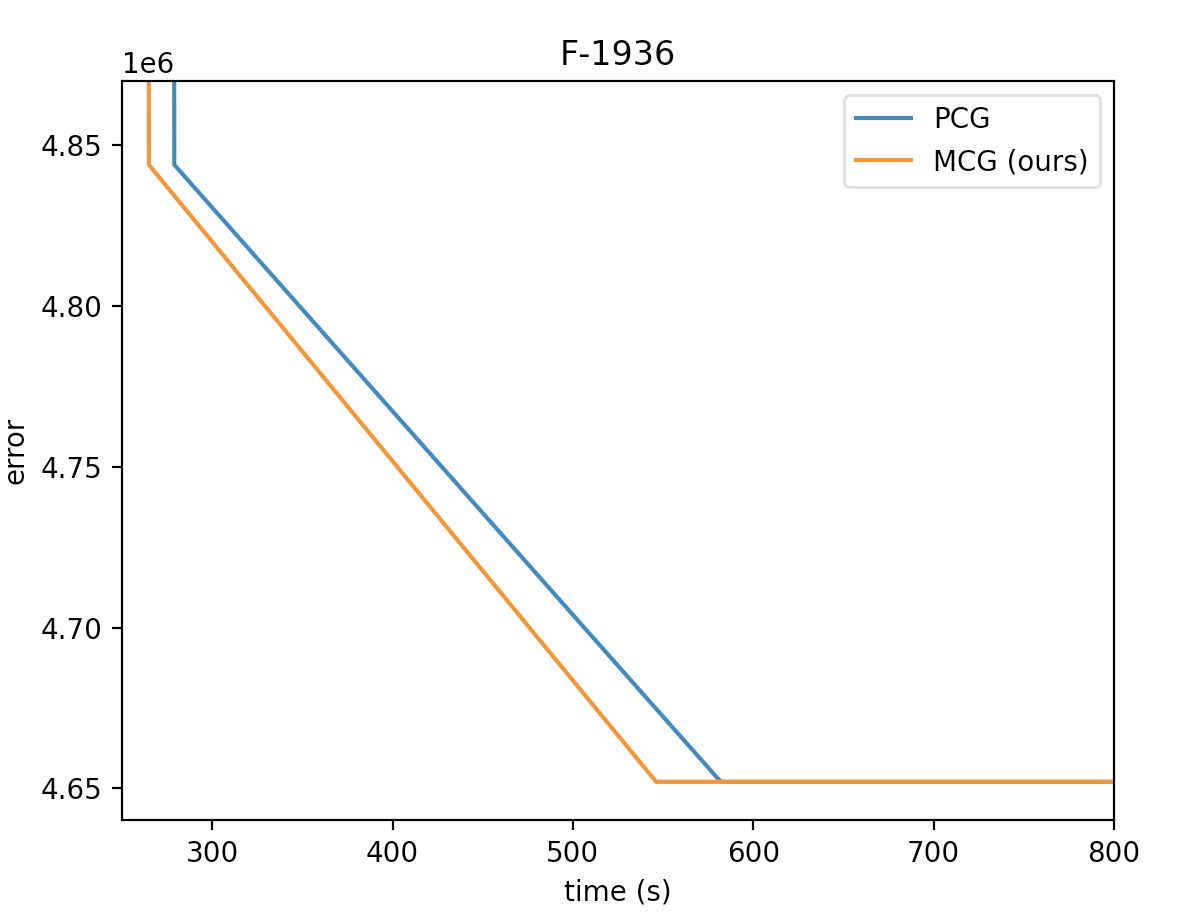} 
\caption{\textit{Final-1936} from BAL dataset}
\label{fig:subim1}
\end{subfigure}
\begin{subfigure}{0.5\textwidth}
\includegraphics[width=0.90\linewidth, height=5cm]{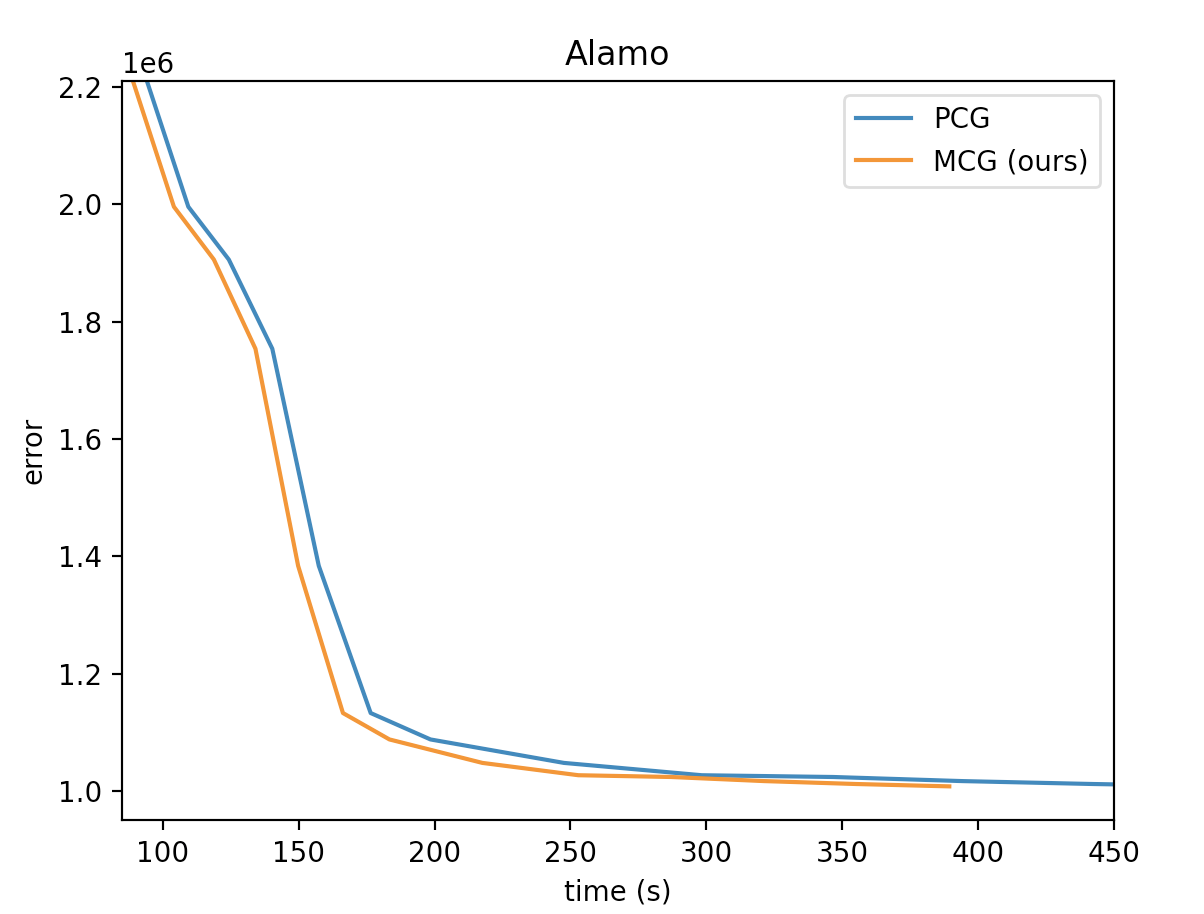}
\caption{\textit{Alamo} from 1dSfM dataset}
\label{fig:subim2}
\end{subfigure}
\caption{Convergence plots of (a) \textit{Final-1936} from BAL dataset and (b) \textit{Alamo} from 1dSfm dataset. The \textit{y}-axes show the total BA cost.}
\label{fig:image2}
\end{figure}

\begin{table}
\caption{Relative performances of MCG w.r.t. PCG. $d$ is the density of the associated Schur complement matrix. MCG greatly outperforms PCG (up to 61\% faster) for dense Schur matrix and competes PCG for sparse Schur matrix. The global BA resolution is up to 22\% faster.} \label{tab1} \centering
\begin{tabular}{|c|c|c|c|}
\hline
Name &  Solver runtime ratio & Global runtime ratio & $d$\\
\hline
Notre-Dame & \textbf{0.39} & \textbf{0.86} & 0.77\\
Alamo & \textbf{0.44} & \textbf{0.78} & 0.77\\
F-394 & \textbf{0.55} & \textbf{0.80} & 0.94\\
F-1936 & \textbf{0.61} & \textbf{0.84} & 0.97 \\
Piazza del Popolo & \textbf{0.66} & \textbf{0.87} & 0.57 \\
Montreal & \textbf{0.77} & 0.96 & 0.60\\
Metropolis & \textbf{0.84} & 0.92 & 0.50 \\
F-871 & 0.93 & 0.97 & 0.40\\
L-646 & 1.01 & 1.00 & 0.25\\
\hline
\end{tabular}
\end{table}

\section{Conclusion}
We propose a novel iterative solver that accelerates the solution of the normal equation for large-scale BA problems. The proposed approach generalizes the traditional preconditioned conjugate gradients algorithm by enlarging its search-space leading to a convergence in much fewer iterations. Experimental validation on a multitude of large scale BA problems confirms a significant speedup in solving the normal equation of up to 61\%, especially for dense Schur matrices where baseline techniques notoriously struggle.  Moreover, detailed ablation studies demonstrate the robustness to variations in the hyper-parameters and increasing speedup as a function of the problem density.

\end{document}